\documentclass{article} 
\usepackage{collas2023_conference,times}
\usepackage{easyReview}

\usepackage{hyperref}
\hypersetup{
    colorlinks=true,
    linkcolor=red,
    filecolor=magenta,
    urlcolor=blue,
    citecolor=purple,
    pdftitle={Overleaf Example},
    pdfpagemode=FullScreen,
    }

\usepackage{microtype}
\usepackage{graphicx}
\usepackage{booktabs} 

\usepackage{amsmath}
\usepackage{amssymb}
\usepackage{mathtools}
\usepackage{amsthm}
\usepackage{comment}
\usepackage{tablefootnote}
\usepackage{subcaption}
\usepackage{multirow}
\usepackage{diagbox}
\usepackage{slashbox}

\usepackage{hyperref}
\hypersetup{
    colorlinks=true,
    linkcolor=red,
    filecolor=magenta,
    urlcolor=blue,
    citecolor=purple,
    pdftitle={Overleaf Example},
    pdfpagemode=FullScreen,
    }

\newcommand{\climb}{CLiMB}
\newcommand{\climbfull}{Continual Learning in Multimodality Benchmark}
\newcommand{\vilt}{ViLT}

\newcommand{\model}{\mathcal{M}}
\newcommand{\adapter}{\mathcal{A}}
\newcommand{\fusion}{\mathcal{F}}
\newcommand{\task}[1]{\mathcal{T}_{#1}}
\newcommand{\dataset}[1]{\mathcal{D}_{#1}}
\newcommand{\loss}{\mathcal{L}}
\newcommand{\fusiontransfer}[1]{\mathbb{T}_{\mathcal{K}}(#1)}
\newcommand{\avgfusiontransfer}{\Tilde{\mathbb{T}}_{\mathcal{K}}}
\newcommand{\algoshort}{$I2I$}
\newcommand{\algovariant}[1]{$I2I_{_{#1}}$}
\newcommand{\algofull}{\textit{Improvise to Initialize}}
\newcommand{\cti}{\text{ClosestTaskInit}}
\DeclareMathOperator*{\argmax}{arg\,max}
\DeclareMathOperator*{\argmin}{arg\,min}

\title{I2I: Initializing Adapters with Improvised Knowledge}

\author{
  Tejas Srinivasan$^1$, Furong Jia$^1$, Mohammad Rostami$^{1,2}$, Jesse Thomason$^1$ \\
  $^1$University of Southern California \\ $^2$USC Information Sciences Institute \\
\texttt{tejas.srinivasan@usc.edu} \\
}

\collasfinalcopy 

\begin{document}

\maketitle

\begin{abstract}
Adapters present a promising solution to the catastrophic forgetting problem in continual learning. 
However, training independent Adapter modules for every new task misses an opportunity for cross-task knowledge transfer. 
We propose \algofull\ (\algoshort), a continual learning algorithm that initializes Adapters for incoming tasks by distilling knowledge from previously-learned tasks' Adapters.  
We evaluate \algoshort\ on \climb, a multimodal continual learning benchmark, by conducting experiments on sequences of visual question answering tasks. 
Adapters trained with \algoshort\ consistently achieve better task accuracy than independently-trained Adapters, demonstrating that our algorithm facilitates knowledge transfer between task Adapters. 
\algoshort\ also results in better cross-task knowledge transfer than the state-of-the-art AdapterFusion without incurring the associated parametric cost. 
\footnote{Our code is available at \url{https://github.com/GLAMOR-USC/CLiMB/tree/i2i}.}
\end{abstract}

\section{Introduction}
Continual Learning (CL) is a learning setting where a single model must learn incoming tasks sequentially, without access to previous tasks' training data when learning new ones~\citep{chen2018lifelong}. 
CL presents models with the dual challenges of effectively transferring knowledge across tasks while mitigating catastrophic forgetting~\citep{french1999catastrophic}. 
Learning strategies that finetune the full pre-trained model, suffer from catastrophic forgetting in the CL setting---when learning new tasks, previous task parameters get overwritten, resulting in diminished model performance on older tasks. 
Further, finetuning pre-trained models on intermediate tasks can harm the model's ability to generalize to new tasks, as the model's parameters diverge further from the pre-trained model checkpoint~\citep{pruksachatkun2020intermediate}. 
While regularization~\citep{kirkpatrick2017overcoming} and replay~\citep{chaudhry2018efficient} methods can mitigate these issues, none of these are perfect solutions. 
Existing CL algorithms suffer from the forgetting issue while also failing to effectively utilize knowledge from both the pre-trained model and intermediate task checkpoints.

Adapters~\citep{houlsby2019parameter} present a promising solution to the catastrophic forgetting problem for Transformer-based CL models. 
Adapters are bottleneck Multi-Layer Perceptron networks that are trained on tasks by inserting inside a frozen pre-trained Transformer. 
In the CL setting, we can train a separate Adapter module for each task, allowing models to retain previous tasks' knowledge by keeping the shared pre-trained Transformer frozen. 
However, independently training Adapter modules for each new task prevents the model from utilizing previously-learned Adapter knowledge. 
AdapterFusion~\citep{pfeiffer2021adapterfusion} proposes a two-phase algorithm: \textit{knowledge extraction} by first learning an Adapter module for the new task, and \textit{knowledge composition} by fusing knowledge from multiple task Adapters through an Attention layer. 
However, adding an AdapterFusion layer to the model for each task adds a large parametric cost, with $\approx$ 20 -- 40$\%$ parameter increase over the base Transformer for each task-specific AdapterFusion layer added.

We propose \algofull\ (\algoshort), a three-phase CL algorithm that utilizes knowledge from previously-learned task Adapters to learn an initialization for the incoming task's Adapter module. 
We initially \textit{improvise} on the incoming task by learning an AdapterFusion over the previous tasks' Adapters. 
We then \textit{initialize} an Adapter for the new task by distilling knowledge from the AdapterFusion trained in the first phase. 
Finally, we train the initialized Adapter on the new task. 
By discarding the AdapterFusion after knowledge distillation, we can avoid the parametric cost while still fusing knowledge learned from previously-seen tasks to enable cross-task knolwedge transfer.

We perform experiments on \climb~\citep{srinivasan2022climb}, a multimodal CL framework, by training models on sequences of visual question answering tasks. 
\algoshort\ facilitates knowledge transfer between task Adapters, outperforming AdapterFusion on learning new tasks \textit{without incurring the large parametric cost}. 
To mitigate \algoshort's training time cost, we experiment with variants that do not require the full training data for the \textit{Improvise} and \textit{Initialize} phases. 
These variants reduce the training overhead while outperforming independently-trained Adapters and AdapterFusion.

\pagebreak

The primary contributions of this work are as follows:
\begin{itemize}
    \item We  propose \algoshort: \algofull, an Adapter-based continual learning algorithm that initializes new task Adapters by first improvising using existing Adapter knowledge.
    \item We perform experiments on sequences of visual question answering tasks in the \climb\ framework, showing that \algoshort\ outperforms AdapterFusion without requiring additional model parameters for knowledge transfer.
    \item We analyze each phase of our \algoshort\ algorithm, and show that improvising from existing Adapters results in better performance on the incoming task.
\end{itemize}

\section{Related Work}
\begin{figure*}
    \centering
    \includegraphics[width=0.75\linewidth]{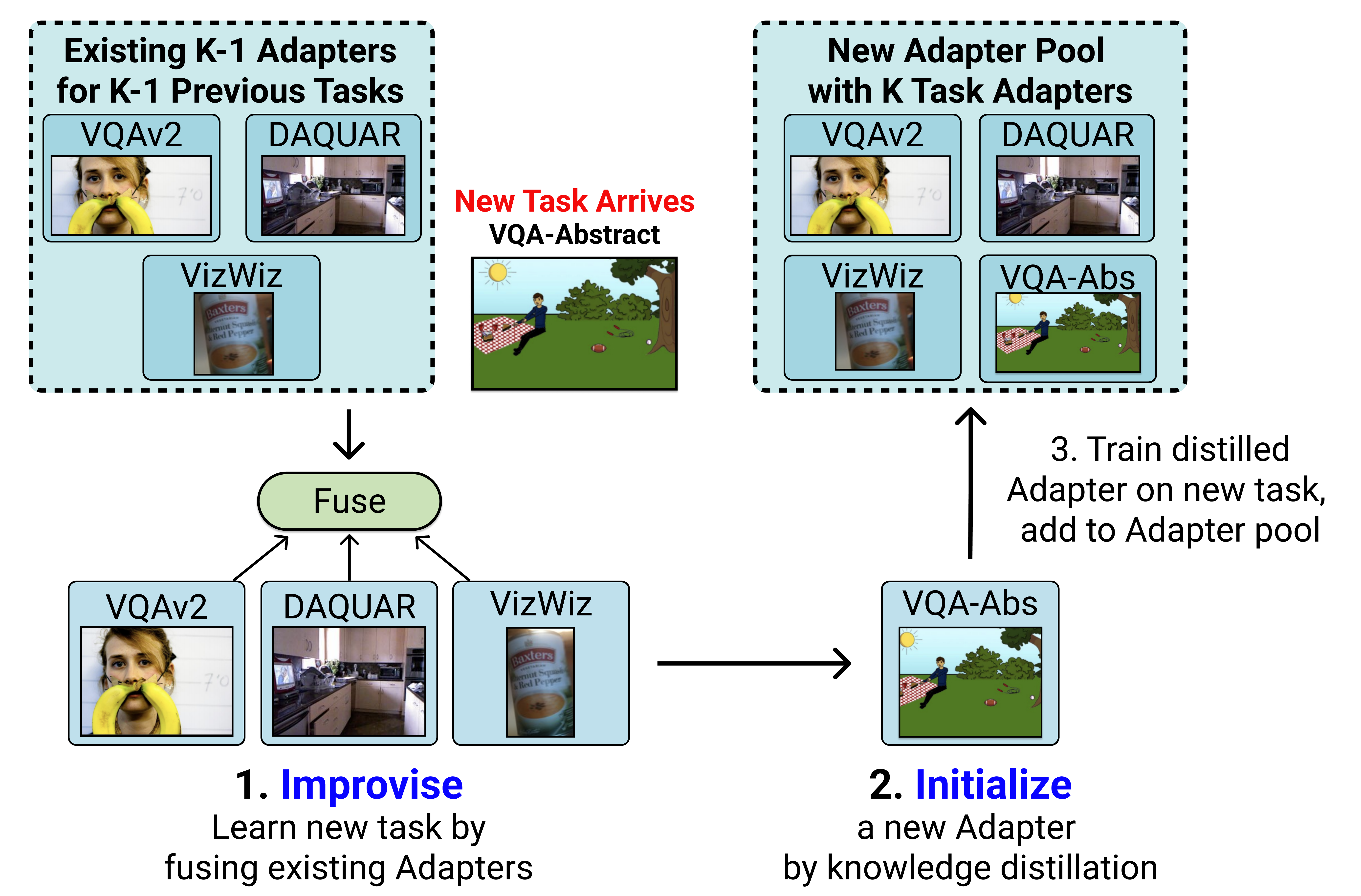}
    \caption{We propose \algoshort: \algofull, an Adapter-based continual learning algorithm that initializes new task Adapters by first improvising using existing Adapter knowledge.}
    \label{fig:overview}
\end{figure*}

We apply \algoshort\ in a sequential, multimodal task learning setting and use Adapters as the basis for knowledge transfer.

\paragraph{Multimodal Continual Learning} Task-incremental continual learning has been primarily studied in unimodal settings~\citep{rebuffi2017learning, mccann2018natural}. 
Within multimodal learning, various works have explored CL over visual question answering, where CL happens over question types~\citep{greco2019psycholinguistics}, visual scenes~\citep{lei2022symbolic} and distinct VQA tasks~\citep{zhang2022cl}. 
\climb~\citep{srinivasan2022climb} constructs a more general framework for continual learning over vision-and-language tasks. 
\citet{suhr2022continual} move beyond vision-and-language into interactive embodied environments, where an instruction-following agent must continually learn from user feedback. 
These works primarily rely on traditional CL methods such as weight consolidation~\citep{kirkpatrick2017overcoming} or experience replay~\citep{chaudhry2018efficient} to mitigate forgetting which either compromise learning capacity of the model or require a memory buffer. 
Our focus is on adoption of Adapters within Transformer models in a CL setting. 

\paragraph{Transfer Learning with Adapters}
Adapters~\citep{houlsby2019parameter} are task-specific modules inserted within a frozen pre-trained Transformer and enable parameter-efficient fine-tuning. 
The size of adapters is typically significantly less than the pre-trained Transformer.
Applied to continual learning, Adapters can solve the catastrophic forgetting issue by learning a separate set of Adapter parameters for each task. 
However, training independent Adapter modules for each task prevents knowledge transfer between the task Adapters. 
One approach to remedy this issue is weight sharing between Adapters. 
~\citet{sung2022vl} experiment with \textit{Half-Shared Adapters}, where different task Adapters shared the same weights for the upsampling layer.
\citet{zhang2022continual} first identify which Adapter layers from previous tasks can be re-used for the new task, and then learn parameters for the remaining Adapter layers for the new task. 
\citet{jin2021learn} train a Hypernetwork~\citep{von2020continual} to generate Adapter weights for different tasks---tasks with similar representations would generate similar Adapter parameters from the Hypernetwork. 
AdapterFusion~\citep{pfeiffer2021adapterfusion} differs from these parameter sharing approaches, by combining representations from multiple task Adapters to improve performance on a target task. 
Our \algoshort\ algorithm is closest in spirit to AdapterFusion, but uses knowledge fusion to learn an initialization for the new task Adapter rather than as a post-hoc transfer learning step.

\section{Methodology}

We propose \algoshort: \algofull, an Adapter-based algorithm that leverages knowledge from already-learned task Adapters when creating Adapters for a new task.

\subsection{Preliminaries}
\label{subsec:prelims}

We describe the continual learning problem (Section~\ref{subsubsec:cl}), and how Adapters (Section~\ref{subsubsec:adapters}) solve the catastrophic forgetting problem. 
AdapterFusion (Section~\ref{subsubsec:adafus}) seeks to solve the lack of cross-task knowledge transfer with traditional vanilla Adapters. 

\subsubsection{Continual Learning}
\label{subsubsec:cl}

We consider a \textit{task-incremental} continual learning setting, where a model encounters a sequence of $K$ distinct tasks, $\task{1...K}$, in order. 
In this work, the initial model is a pre-trained Transformer $\model_0$. 
For every task $\task{i}$, the model is initialized with the previous checkpoint $\model_{k-1}$, and trained to minimize training loss on the dataset $\dataset{i}$:
\begin{align}
    \model_k \leftarrow \argmin_{\model} \loss(\dataset{k}; \model).
\end{align}
After learning task $\task{i}$, the model cannot access the training data $\dataset{k}$ or the previous model checkpoints $\model_{0...k-1}$. 
We consider a \textit{task-aware} continual learning setting, where at test time we know the task identity for every model input.

\subsubsection{Adapters}
\label{subsubsec:adapters}

Adapter modules~\citep{houlsby2019parameter} are Multi-Layer Perceptron layers typically inserted within each layer of the pre-trained Transformer.
In general, they amount to $\approx 1\%$ of the Transformer model $\model$ parameters.
When training on task $\task{}$, the Transformer parameters $\model_0$ are kept frozen while the Adapter modules $\Phi$ are learned. 
Some additional task-specific parameters $\Psi$ outside the Transformer may also be learned, such as a classification head for discriminative models. 
In our model, $\Psi$ is a linear layer that projects visual features before passing into a language model (Figure~\ref{fig:clipbart}).

In the continual learning setting, we learn Adapter parameters $\Phi_{k}$ and task-specific parameters $\Psi_k$ for every task $\task{k}$:
\begin{align}
    \Phi_k , \Psi_k \leftarrow \argmin_{\Phi, \Psi} \loss (\dataset{k} ; \model_0, \Phi, \Psi).
\end{align}
When learning task $\task{k}$, the previous tasks' Adapter modules $\Phi_{1...k-1}$ remain untouched. 
Since we are operating in a task-aware setting, at inference time we can load the Adapter modules associated with the corresponding inference time task.
In that way, even after learning $K$ tasks, the model retains the same performance on the evaluation set of $\task{k}$ as when the Adapter module $\Phi_k$ was originally trained. 
In other words, there is no catastrophic forgetting during the continual learning at the cost of adding $\approx K \times 1\%$ Adapter parameters to the base model.

However, by training Adapters independently on distinct datasets, there is no cross-task knowledge transfer. 
We hypothesize that Adapters can benefit knowledge acquired while learning previous tasks to achieve the best of both worlds: forward knowledge transfer without forgetting past tasks during continual learning.

\subsubsection{AdapterFusion}
\label{subsubsec:adafus}

AdapterFusion~\citep{pfeiffer2021adapterfusion} is a two-phase transfer learning method that composes knowledge from multiple task Adapters to improve model performance on individual tasks. 
In the first \textit{knowledge extraction} phase, the model learns an Adapter $\Phi_k$ for task $\task{k}$. 
In the second \textit{knowledge composition} phase, the model is again trained on task $\task{k}$ using a Fusion layer $\fusion_k$ that combines representations from multiple frozen task Adapters $\Phi_{1...k}$. 
The Fusion layer $\fusion_{k}$ is typically an Attention layer inserted after the Adapter modules $\Phi$ within each Transformer layer.
\begin{align}
    \text{Phase 1: }\Phi_k , \Psi_k & \leftarrow \argmin_{\Phi, \Psi} \loss (\dataset{k} ; \model_0, \Phi, \Psi) \\
    \text{Phase 2: }\fusion_k , \Psi_k & \leftarrow \argmin_{\fusion, \Psi} \loss (\dataset{k} ; \model_0, \fusion(\Phi_1...\Phi_k), \Psi) \\
    \fusion(\Phi_1...\Phi_k) & = \texttt{Attn}\big(Q=x, \quad K,V=\Phi_1(x)...\Phi_k(x)\big)
\end{align}
AdapterFusion facilitates forward transfer learning between different task Adapters, but comes with an exponential cost in parameters with respect to the number of tasks to be learned.
Since the Fusion layer consists of Query, Key, and Value matrices, the added parameters from a single task's Fusion layer range from 20-40\% of the pre-trained Transformer parameters. 
As the number of tasks $K$ increases, this added parameter cost will even exceed the pre-trained Transformer size. 
For instance, with a pre-trained \vilt\ model, the added AdapterFusion layers will have more parameters than the original \vilt\ Transformer after learning just 5 tasks in the CL setting considered here.

Additionally, performing the fusion operation as a post-hoc transfer learning step may be sub-optimal. 
Specifically, after the model has already learned an Adapter $\Phi_k$ that has converged to a local minima, it may be difficult to encourage the model to move out of that local minima using knowledge from Adapters $\Phi_{1...k-1}$.

\begin{figure}
    \centering
    \includegraphics[width=\linewidth]{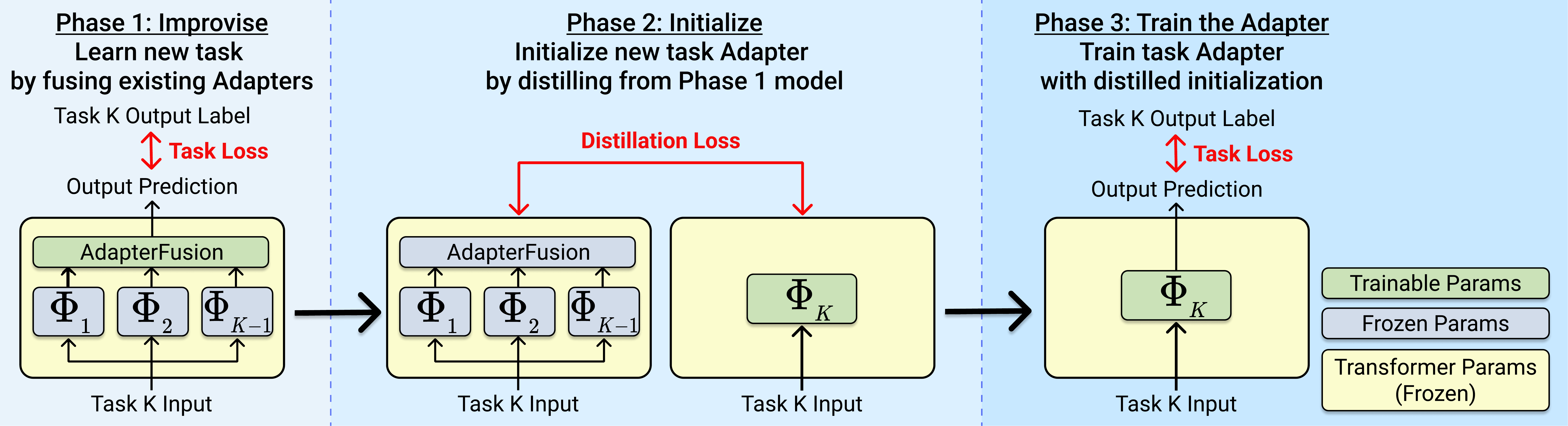}
    \caption{
        Our proposed \algofull\ algorithm. 
        We first \textit{improvise} on a new task $\task{k}$ by training a Fusion layer $\fusion_{k}(\Phi_1, \Phi_2, ..., \Phi_{k-1})$ that learns to fuse representations from each of the previously-learned Adapters $\Phi_{1...k-1}$. 
        We then \textit{initialize} a new Adapter $\Phi_{k}$ by distilling knowledge from the Fusion layer, and discard the parameters of the much larger Fusion layer.
        Finally, the distilled Adapter $\Phi_{k}$ is trained on the task $\task{k}$.
    }
    \label{fig:method}
\end{figure}

\subsection{Improvise to Initialize}
\label{subsec:i2i}

We hypothesize that fusing knowledge from existing tasks $\task{1...k-1}$ to learn an initialization for a new task Adapter $\Phi_k$ will yield better knowledge transfer than AdapterFusion's post-hoc knowledge composition.
We propose \algoshort: \algofull\ (Figure~\ref{fig:method}), a three-phase training strategy that initially learns a fusion of already-learned Adapters $\Phi_{1...k-1}$, initializes a new task Adapter $\Phi_{k}$ by distilling knowledge from the fusion, and then trains the Adapter $\Phi_k$ as usual. 
We apply the \algoshort\ algorithm to learn Adapters $\Phi_k$ for $k > 1$. 
For task $\task{1}$, we directly train Adapter $\Phi_1$ on training data $\dataset{1}$.

\paragraph{Phase One: Improvise} We first \textit{improvise} on the new task, using knowledge from the already-trained Adapters. 

When learning the second task, \textit{i.e.} $k = 2$, we only have one already-learned Adapter $\Phi_1$. 
We minimize the training loss $\loss$ on the dataset $\dataset{k}$ by learning task-specific parameters $\Psi_2$ using the frozen Adapter $\Phi_1$.
\begin{align}
    \Psi_2 \leftarrow \argmin_{\Psi} \loss(\dataset{k} ; \model_0, \Phi_1, \Psi)
\end{align}

When combining multiple already-learned Adapters, i.e. $k \geq 3$, we additionally train a Fusion layer $\fusion_{k}(\Phi_1...\Phi_{k-1})$.
\begin{align}
    \fusion_{k}, \Psi_k \leftarrow \argmin_{\fusion, \Psi} \loss(\dataset{k} ; \model_0, \fusion(\Phi_1...\Phi_{k-1}), \Psi)
\end{align}
The only parameters trained in the \textit{Improvise} phase are the Fusion parameters $\fusion_k$ and the task-specific parameters $\Psi_k$.

\paragraph{Phase Two: Initialize} In Phase Two, we \textit{initialize} a new Adapter $\Phi_k$ by distilling knowledge from the model trained in the \textit{Improvise} phase. 

For the second task $\task{2}$, we can initialize the new Adapter $\Phi_2$ by directly copying the parameters of Adapter $\Phi_1$, and copying the task-specific parameters $\Psi_2$ learned from the \textit{Improvise} phase. 

When initializing the Adapter $\Phi_k$ for $k \geq 3$, we use knowledge distillation. 
The teacher model, $\model_T$, is the previously-learned model with the Fusion layer $\fusion_{k}(\Phi_1...\Phi_{k-1})$. 
The student model, $\model_S$, is a new Adapter $\Phi_{k}$ inserted inside the pre-trained Transformer $\model_0$. 
The task-specific parameters $\Psi_k$ for the student model are copied from the teacher model.

During distillation, the student model $\model_S$ is trained to produce the same representations as the frozen teacher model $\model_T$, by minimizing a distillation loss $\loss_D$. 
The only parameters trained during this phase are the student model's task-specific parameters $\Psi_k$ and Adapter $\Phi_k$.
\begin{align}
    \Phi_{k}, \Psi_K \leftarrow \argmin_{\Phi, \Psi} \loss_{D} (\model_T(\mathbf{x}) , \model_S(\mathbf{x}))
\end{align}
After completing the distillation phase, we can discard the Fusion layer $\fusion_{k}$, alleviating our model from the parametric growth that AdapterFusion suffers from.

\paragraph{Phase Three: Train the Adapter} The Adapter $\Phi_k$ is again trained on task $\task{k}$, using the Adapter $\Phi_k$ and task-specific parameters $\Psi_k$ from the Phase Two student model as the initial checkpoint. 
\begin{align}
    \Phi_k, \Psi_k \leftarrow \argmin_{\Phi} \loss (\dataset{k} ; \model_0, \Phi, \Psi)
\end{align}
Since we discard the Fusion layer $\fusion_k$, the only added parameters for each task are from the Adapter $\Phi_{k}$, which are typically $\approx 1\%$ of the full Transformer size. 
Our \algoshort\ algorithm solves the large parametric cost of AdapterFusion, while achieving cross-task knowledge transfer.

\subsubsection{Mitigating I2I's Training Time Cost}

The \algoshort\ algorithm performs three passes over the training data, making our training procedure more time-consuming than standard Adapter training. 
We propose three variants of our algorithm to mitigate the training time cost:
\begin{enumerate}
    \item \algovariant{FF}: The model is trained using the Full training data in both the \textit{Impovise} and \textit{Initialize} phases.
    \item \algovariant{FL}: The model is trained using the Full training data for the \textit{Improvise} phase, but the \textit{Initialize} phase is trained using a Low-shot version of the training data ($5\%$ in our experiments).
    \item \algovariant{LL}: The \textit{Improvise} as well as \textit{Initialize} phases are trained using Low-shot versions of the training data.
\end{enumerate}
In all three variants, Phase Three of the algorithm, \textit{i.e.} final training of the Adapter $\Phi_k$ on the task data $\dataset{k}$, is performed using the full training data.

\section{Experiments}
We perform Continual Learning over sequences of visual question answering tasks (Section~\ref{subsec:tasks}) on a CLIP-BART model (Section~\ref{subsec:model}). 
We compare our method against the AdapterFusion baseline (Section~\ref{subsec:baseline}), and evaluate each algorithm's ability to transfer knowledge across Adapters (Section~\ref{subsec:metrics}).

\subsection{Continual Learning Tasks}
\label{subsec:tasks}

\begin{table}[t]
  \centering
  \begin{tabular}{llll}
    \bf Task & \bf Image source & \bf Train/Val QA Pairs & \bf Score Metric \\
    \toprule
    VQAv2 & MSCOCO images  & 443k/214k & VQAScore\tablefootnote{https://visualqa.org/evaluation.html} \\
    Visual7W & MSCOCO images & 69.8k/28k & Exact Answer Match \\
    VQA-Abstract & Abstract scenes & 60k/30k & VQAScore \\
    VizWiz & Images captured by blind people & 20k/4.3k & VQAScore \\
    DAQUAR & Indoor household scenes from NYU Depth V2  & 10k/2.5k & Exact Answer Match \\
    \bottomrule
  \end{tabular}
  \caption{We experiment with continual learning over five visual question answering tasks.}
  \label{tab:tasks}
\end{table}

We perform experiments on \climb, the \climbfull.  
The \climb\ benchmark contains a variety of vision-language tasks, including question answering, visual entailment, and vision-language reasoning.
We hypothesize that cross-task knowledge transfer is more feasible between similar tasks. 
For this reason, we extend the \climb\ framework and perform experiments on five visual question answering (VQA) tasks (Table~\ref{tab:tasks}):
\textbf{VQAv2}~\citep{balanced_vqa_v2}, \textbf{Visual7W}~\citep{visual7w}, \textbf{VQA-Abstract}~\citep{antol2015vqa}, \textbf{VizWiz}~\citep{gurari2018vizwiz}, and \textbf{DAQUAR}~\citep{daquar}. 
We evaluate continual learning algorithms on three randomly selected task orders: 
\begin{enumerate}
    \itemsep0em
    \item VQAv2 $\rightarrow$ Visual7W $\rightarrow$ VQA-Abstract $\rightarrow$ DAQUAR $\rightarrow$ VizWiz
    \item VizWiz $\rightarrow$ DAQUAR $\rightarrow$ Visual7W $\rightarrow$ VQA-Abstract $\rightarrow$ VQAv2
    \item DAQUAR $\rightarrow$ VQAv2 $\rightarrow$ VizWiz $\rightarrow$ Visual7W $\rightarrow$ VQA-Abstract
\end{enumerate}

\begin{figure}[t]
    \centering
    \includegraphics[width=0.8\linewidth]{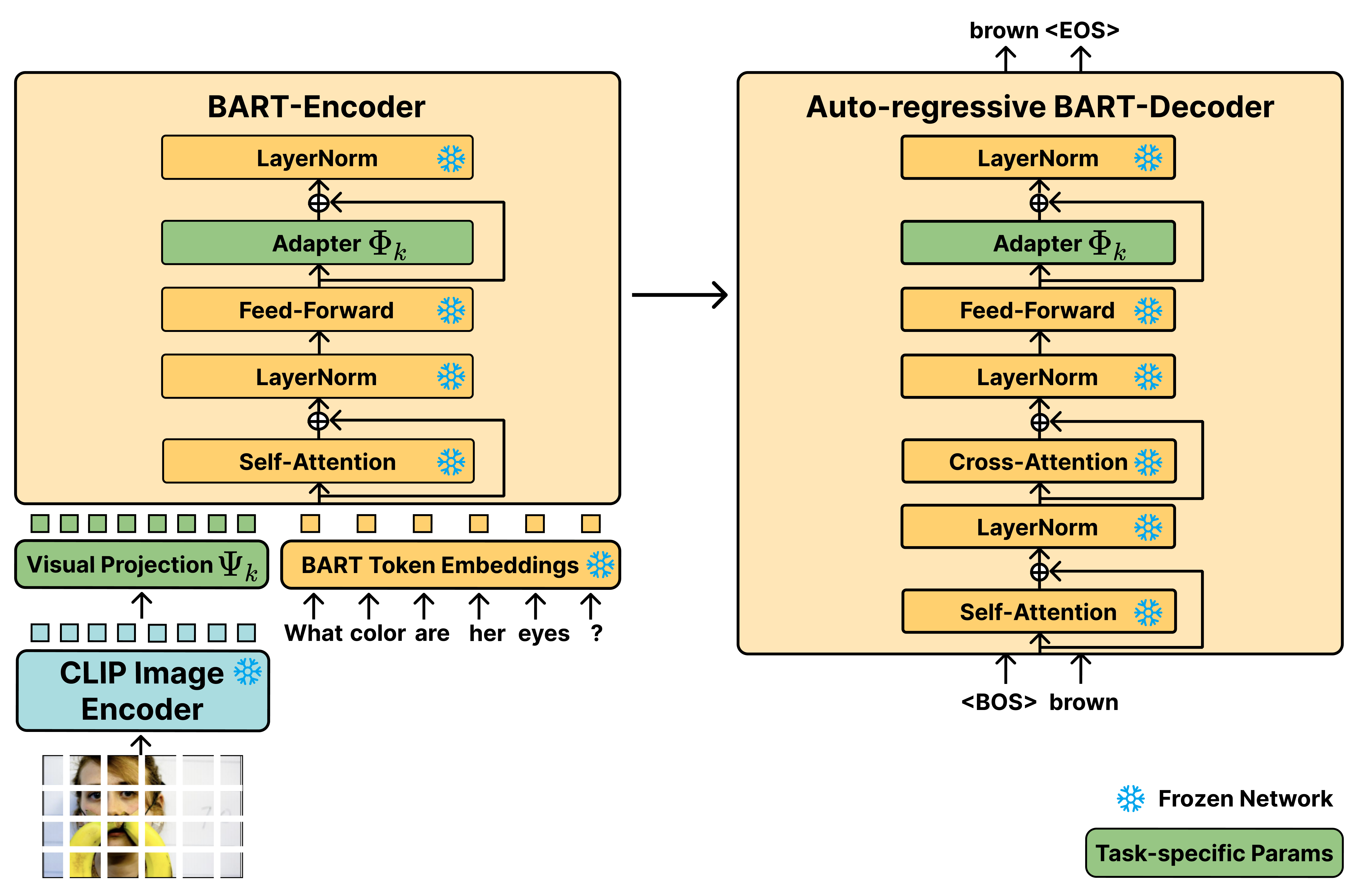}
    \caption{
        The CLIP-BART architecture. 
        The CLIP image encoder is used to extract image representations, and the BART model uses projected visual features and question token embeddings to generate an answer. 
        For training on task $\task{k}$, task-specific parameters $\Psi_k$ and $\Phi_k$ are learned while the CLIP and BART parameters are kept frozen.
        }
    \label{fig:clipbart}
\end{figure}
\subsection{Continual Learning Model}
\label{subsec:model}

We perform Continual Learning over open-ended QA tasks using CLIP-BART, a generative Vision-Language model. 

\paragraph{CLIP-BART Architecture} Following previous work~\citep{sung2022vl}, our CLIP-BART model (Figure~\ref{fig:clipbart}) combines visual representations from CLIP~\citep{radford2021learning} into a text generation BART model~\citep{lewis2020bart}. 
Images are encoded using a frozen CLIP-ViT image encoder~\citep{radford2021learning} into a set of patch features. 
A visual linear projection $\Psi$ is applied on the patch features, and the projected image features are concatenated to the sequence of BART token embeddings before passing into a pre-trained BART-base model. 
BART~\citep{lewis2020bart} is a pre-trained encoder-decoder language model. 
We modify the BART encoder Transformer to jointly encode the image patch features and input token embeddings.
The BART Transformer decoder is trained to generate a sequence of tokens corresponding to the answer, by attending over the encoder features. 
The CLIP-BART model has a total of 227M parameters, with just 139M learnable parameters since the CLIP image encoder is frozen.

\begin{table*}[t]
    \centering
    \begin{tabular}{lrrrrr}
    \bf Method & \multicolumn{1}{c}{\bf VQAv2} & \multicolumn{1}{c}{\bf Visual7W} & \multicolumn{1}{c}{\bf VQA-Abstract} & \multicolumn{1}{c}{\bf VizWiz} & \multicolumn{1}{c}{\bf DAQUAR}  \\
    \toprule
    Full-Model Finetuning    & $64.44$ &	$25.21$ &	$67.55$ &	$43.57$ &	$25.65$ \\
    Independently-Trained Adapters  & $61.42$ &	$24.56$ &	$63.38$ &	$42.04$ &	$23.58$  \\
    \bottomrule
    \end{tabular}
    \caption{We report performance of CLIP-BART trained individually on each VQA task, when the full model is fine-tuned and when only Adapter modules are learned. 
    Note that this single-task learning setting is not comparable with the CL setting in Table~\ref{tab:fusion-transfer} and represents an upper bound on individual task accuracy without knowledge transfer.}
    \label{tab:adapters}
\end{table*}

\paragraph{CLIP-BART Pre-training} Since CLIP-BART is a combination of two separately pre-trained models, we introduce a visual projection layer from CLIP features to BART token inputs tasked with aligning the two pre-trained models for each task. 
We pre-train CLIP-BART on the MS-COCO training dataset using masked language modeling and image-text matching objectives, keeping the CLIP encoder frozen and pre-training only the visual projection layer and the BART model. 
At the start of continual learning, the model's BART parameters are initialized with the pre-trained checkpoint and remain frozen during continual learning. 
Additionally, the pre-trained visual projection layer parameters are used to initialize the task-specific visual projection layer $\Psi_k$ for each task $\task{k}$.

\paragraph{Training CLIP-BART with Adapters} Following~\cite{pmlr-v97-stickland19a}, we insert Adapter modules after the feed-forward network in every Transformer layer in the BART encoder and decoder---Adapters are not added inside the frozen CLIP image encoder. 
The Adapter parameters amount to 0.9M parameters per task, which is less than 0.5\% of the full CLIP-BART model. 
We also train a visual projection layer $\Psi_k$ for each task $\task{k}$. 
In Table~\ref{tab:adapters}, we compare CLIP-BART fine-tuning with Adapter training on our VQA tasks. 

\paragraph{Training CLIP-BART with \textit{I2I}} In \textit{Phase One: Improvise} and \textit{Phase Three: Train the Adapter}, we directly optimize the model using the ground-truth answer as supervision, using a cross-entropy loss over the output tokens. In \textit{Phase Two: Initialize}, we train a student model $\model_S$ to produce the same representations as a teacher model $\model_T$ by minimizing a distillation loss $\loss_D$. 
For CLIP-BART, we minimize the MSE loss between the teacher and student models' encoder hidden states $h^E$ and decoder hidden states $h^D$.
\begin{align}
    h^E_T, h^D_T &\leftarrow \model_T(\mathbf{x}) \\
    h^E_S, h^D_S &\leftarrow \model_S(\mathbf{x}) \\
    \loss_D(\model_T(\mathbf{x}), \model_S(\mathbf{x})) & = MSE(h^E_T, h^E_S) + MSE(h^D_T, h^D_S)
\end{align}

\begin{figure}
    \centering
    \includegraphics[width=0.9\linewidth]{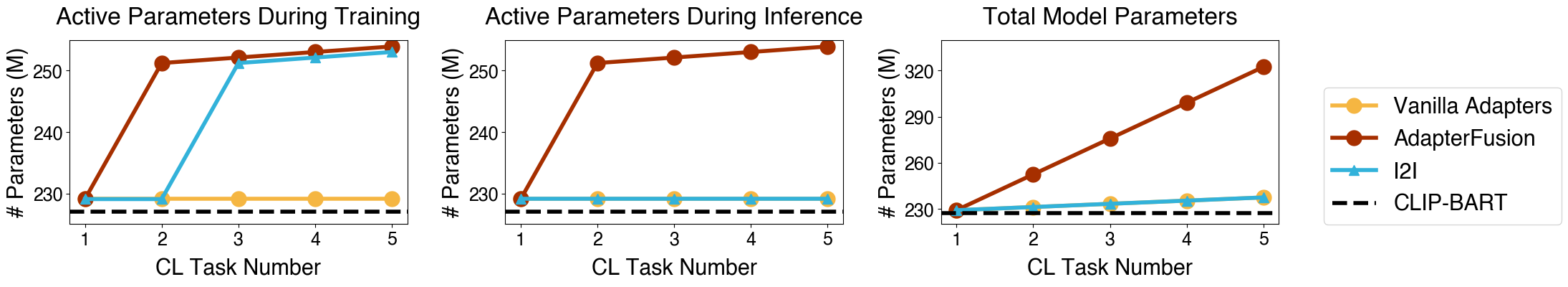}
    \caption{How the number of parameters active during a forward pass during training (left) and inference (center), as well as overall model size (right), increase as each new CL task arrives.
    Note the backbone CLIP-BART model has 227M parameters.}
    \label{fig:params_growth}
\end{figure}

\subsection{Baseline Methods}
\label{subsec:baseline}

In this work, we do not focus on the catastrophic forgetting problem, and instead aim to facilitate cross-task transfer between Adapters. 
Therefore, the only CL algorithms that we compare \algoshort\ against are independently-trained, \textit{vanilla} Adapters, and AdapterFusion. 
Figure~\ref{fig:params_growth} highlights the impact of adding Adapter parameters for each CL algorithm. 
We observe that \algoshort\ uses a similar number of parameters to AdapterFusion while training, due to the training of the fusion layer $\fusion_k$ in \textit{Phase One: Improvise}. 
However, during inference the number of model parameters used is much lower, since the fusion layer $\fusion_k$ is discarded and only the Adapter parameters $\Phi_k$ are added. 
Further, the overall model size using \algoshort\ grows at the same rate as vanilla Adapters, and at a much slower rate than AdapterFusion.
Additionally, we introduce a \cti\ baseline, where each new task's Adapter is initialized using parameters of the most similar already-learned task's Adapter. More details of how the most similar task was selected are available in Appendix~\ref{sec:cti}.

We do not compare against other traditional continual learning methods such as Experience Replay~\citep{chaudhry2018efficient} and EWC~\citep{kirkpatrick2017overcoming}, since these primarily target overcoming forgetting. 
Further, these methods have been shown to hurt the model's ability to generalize to new vision-language tasks~\citep{srinivasan2022climb}.

\subsection{Evaluation Metrics}
\label{subsec:metrics}

Our experiments compare learning algorithms on their ability to transfer knowledge between task Adapters. 
We evaluate knowledge transfer by computing relative improvements for each fusion method over vanilla Adapters. 
If $S_{\fusion}^i$ is the score for task $\task{i}$ using fusion algorithm $\fusion$, and $S_{\adapter}^i$ is the score for the same task using an independently-trained vanilla Adapter, the knowledge transfer for that task, $\fusiontransfer{i}$ is computed as 
\begin{align}
    \fusiontransfer{i} = \frac{S_{\fusion}^{i} - S_{\adapter}^i}{S_{\adapter}^i} \times 100\%.
\end{align}

For a sequence of $K$ tasks $\task{1...K}$, the overall knowledge transfer $\avgfusiontransfer$ is calculated over all tasks $\task{2...K}$:
\begin{align}
    \avgfusiontransfer = \frac{\sum_{i=2}^{K} \fusiontransfer{i} }{K-1} = \frac{1}{K-1}\sum_{i=2}^{K} \frac{S_{\fusion}^{i} - S_{\adapter}^i}{S_{\adapter}^i}.
\end{align}
Note that task $\task{1}$ does not involve fusion, so we do not include it in the overall knowledge transfer calculation.

\subsection{Implementation Details} 
We train our continual learning models with a batch size of 64 on a single 48GB NVIDIA RTX A6000 GPU. 
When training on a single sequence of five CL tasks: vanilla Adapters take about 30 hours; AdapterFusion takes 57 hours; and $\algoshort_{LL}, \algoshort_{LF}$ and $\algoshort_{FF}$ take 49, 63 and 83 hours respectively.
Note that at inference time, the \algoshort\ methods runtimes are equivalent to vanilla Adapters (Figure~\ref{fig:params_growth}) while achieving higher performance than the parameter-heavy AdapterFusion method (Table~\ref{tab:fusion-transfer}).

\section{Results and Discussion}

\begin{table*}[t]
    \centering
    \footnotesize
    \begin{tabular}{lrrrrrr}
    & \multicolumn{5}{c}{Individual Task Knowledge Transfer $\fusiontransfer{i}$ and Task Score $[S^i] (\%)$} & \multirow{3}{*}{\shortstack{Overall\\Knowledge\\Transfer, $\avgfusiontransfer$}} \\
    \cmidrule{2-6}
    & & & & & \\
    Method & \multicolumn{1}{c}{VQAv2} & \multicolumn{1}{c}{Visual7W} & \multicolumn{1}{c}{VQA-Abstract} & \multicolumn{1}{c}{VizWiz} & \multicolumn{1}{c}{DAQUAR} &  \\
    \toprule
    Vanilla Adapters & [61.42] &   [24.56] &   [63.38] &   [42.04] &   [23.58] & 0.00\% \\
    \midrule
    AdapterFusion   & 0.16\% [61.52] & -5.40\% [23.23] &   -4.93\% [60.26] &   3.04\% [43.32] & -0.74\% [23.41]    & -2.08\%      \\
    ClosestTaskInit & -0.22\% [61.28] & 1.05\% [24.82] & 2.90\% [65.22] & 1.71\% [42.76] & 1.90\% [24.03] & 1.90\% \\ 
    \midrule
    \algovariant{LL} &  -0.16\% [61.32] &    2.78\% [25.24] & -0.72\% [62.92] & 2.97\% [43.29]  & -0.55\% [23.45] & 1.17\%  \\
    \algovariant{FL}  & 0.32\% [61.62] &    2.71\% [25.23] & 0.66\% [63.80] & \pmb{4.42\% [43.90]} & -0.68\% [23.42] & 1.94\%   \\
    \algovariant{FF}  & 0.08\% [61.47] & \pmb{4.22\% [25.60]} & \pmb{2.96\% [65.26]} & 3.86\% [43.66] & \pmb{4.65\% [24.68]} & \pmb{4.03\%}  \\
    \bottomrule
    \end{tabular}
    \caption{For each CL algorithm, we report the knowledge transfer $\fusiontransfer{i}$ and task score $[S^i] (\%)$ for every task $\task{i}$, averaged across three task orders. We also report overall knowledge transfer $\avgfusiontransfer$, averaged across three task orders.}
    \label{tab:fusion-transfer}
\end{table*}

In Table~\ref{tab:fusion-transfer}, we report the knowledge transfer $\fusiontransfer{k}$ and task score $S^i$ during continual learning for each task $\task{i}$, averaged across three task orders.
We also report the overall knowledge transfer $\avgfusiontransfer$, averaged across all task orders.

All variants of our \algoshort\ algorithm achieve positive knowledge transfer compared to independently-trained Adapters. 
Increasing the access to training data during the \textit{Improvise} and \textit{Initialize} phases result in better Adapter initialization and subsequent knowledge transfer. 
The \algovariant{FF} variant, which uses the full training data of each task to fuse existing task Adapters and distill to a new one, achieves an overall knowledge transfer of 4\% over vanilla Adapters, averaged over the three task orders. 

We also see that our method results in much better knowledge transfer than AdapterFusion. 
In fact, on average, AdapterFusion results in worse knowledge transfer than independently-trained Adapters. 
Further, we observe that the ClosestTaskInit baseline also achieves positive knowledge transfer, although not as significantly as the \algovariant{FF} variant. 
These results lend credence to our hypothesis that fusing knowledge from other Adapters is more useful for Adapter initialization, rather than as a post-hoc transfer learning step. 

We also see that certain tasks are able to effectively utilize knowledge from previously-learned Adapters than others. 
Visual7W and VizWiz in particular show positive task transfer across all three variants of our method. 
On the other hand, VQAv2 has close to zero knowledge transfer across all methods. 

\section{Analysis}

We systematically analyze each phase of the \algoshort\ algorithm. 
We first evaluate how much new tasks benefit from fusing previous task knowledge in \textit{Phase One: Improvise} (Section~\ref{subsec:analyze-improv}), how much knowledge is transferred to the new Adapter during \textit{Phase Two: Initialize} (Section~\ref{subsec:analyze-init}), and how much each \algoshort\ variant benefits from the final Adapter training in \textit{Phase Three: Train the Adapter} (Section~\ref{subsec:train-analysis}).

\subsection{How much Knowledge is Gained from Improvising?}
\label{subsec:analyze-improv}

We analyze the benefit of fusing existing Adapters in \textit{Phase One: Improvise}. 
In the \textit{Improvise} phase, we train the task-specific visual projection layer $\Psi_k$ and an AdapterFusion of existing Adapters $\fusion_k(\Phi_1...\Phi_{k-1})$. 
For each of our visual question answering tasks, we evaluate the \textit{Improvise} model from our \algovariant{FF} algorithm in all three task orders (omitting those orders in which the task in question appeared first), and average these scores. 

We compare the \textit{Improvise} model with a \textit{Knowledge-Free} baseline. 
We train a CLIP-BART model on each task, freezing the pre-trained CLIP and BART parameters and only training the visual projection layer $\Psi_k$. 
Figure~\ref{fig:improv-analysis} shows that the \textit{Improvise} model results in better task accuracy than the Knowledge-Free baseline, across all tasks. 
These results reveal that the \textit{Improvise} model is able to utilize knowledge from existing task Adapters to learn new tasks better. 

\subsection{How much Knowledge is Lost During Distillation?}
\label{subsec:analyze-init}

In \textit{Phase Two: Initialize}, we distill knowledge from the \textit{Improvise} model to a new task Adapter $\Phi_k$ by minimizing a loss that trains the new Adapter to produce hidden representations similar to that of the \textit{Improvise} model. 
However, this distillation process may result in a performance drop from the teacher model to the student Adapter. 
We measure this performance drop by computing the Distillation Decay for each task $\task{k}$, which is the relative decrease in task acurracy from the \textit{Improvise} model ($S_{\fusion_k}$) to the distilled Adapter ($S_{\Phi_k}$).
\begin{align}
    \text{Distillation Decay for task } \task{k} = \frac{S_{\fusion_k} - S_{\Phi_k}}{S_{\fusion_k}} \times 100\%
\end{align}

\begin{figure}[t]
    \centering
    \begin{subfigure}[t]{.3\textwidth}
      \centering
      \includegraphics[width=0.97\linewidth]{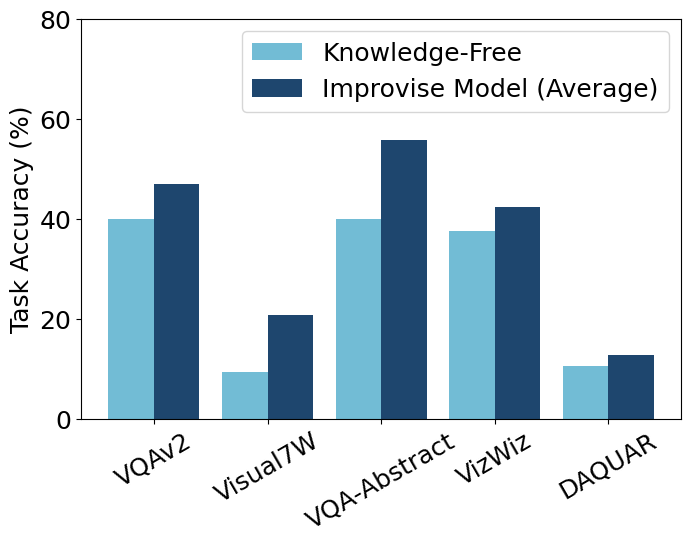}
      \caption{}
      \label{fig:improv-analysis}
    \end{subfigure} \hfill
    \begin{subfigure}[t]{.66\textwidth}
      \centering
      \includegraphics[width=0.96\linewidth]{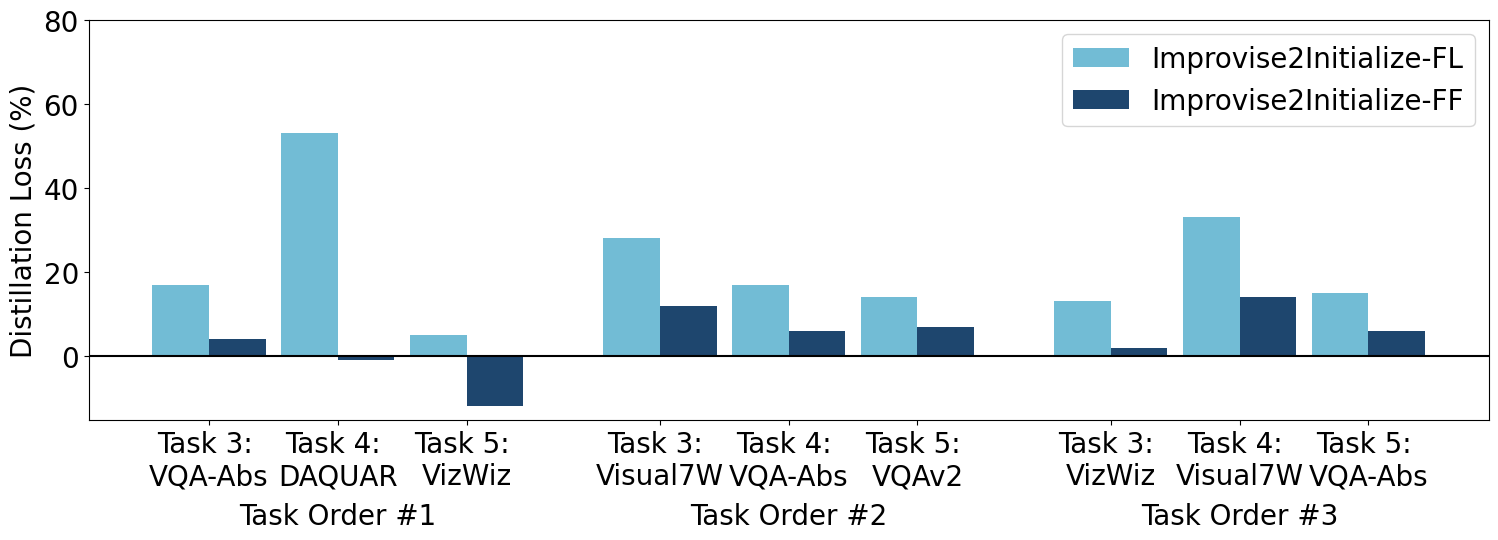}
      \caption{}
      \label{fig:distill-analysis}
    \end{subfigure}
    \begin{subfigure}[t]{.665\textwidth}
      \centering
      \includegraphics[width=0.96\linewidth]{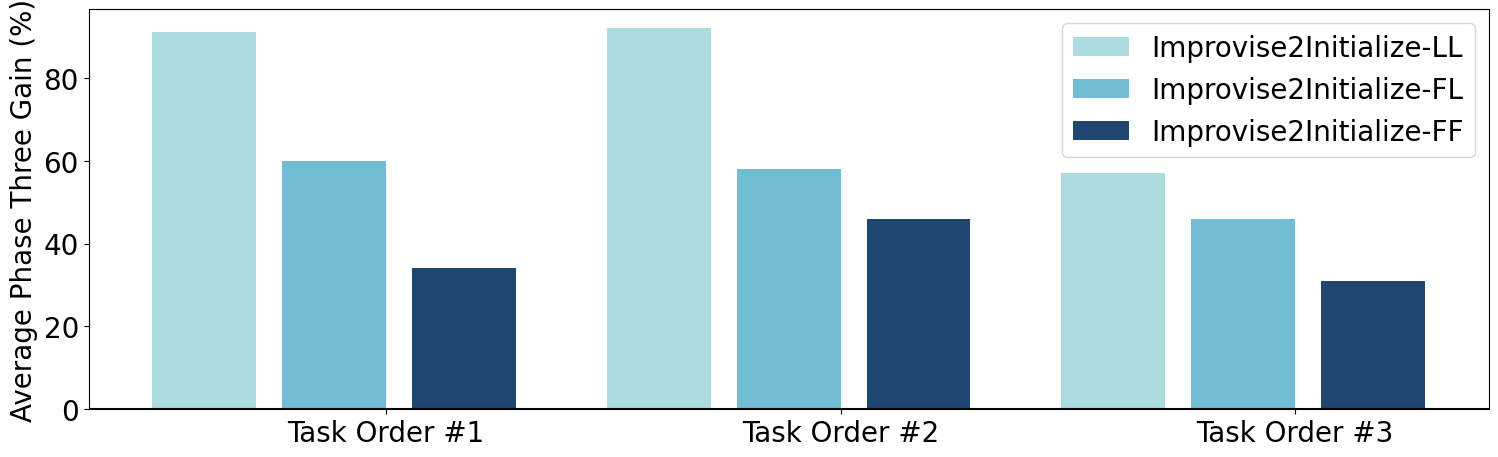}
      \caption{}
      \label{fig:train-analysis}
    \end{subfigure}
    \caption{\textbf{(a)} Task accuracies for the \textit{Knowledge-Free} model and the \textit{Improvise} models (averaged over all task orders) \textbf{(b)} The Distillation Decay 
      for Adapters trained with an \textit{Initialize} phase, using Low-Shot and Full training data (\algovariant{FL} and \algovariant{FF}, respectively). 
      \textbf{(c)} The Average Phase Three Gain for each \algoshort\ variant, for each of the three task orders.}
    \label{fig:analysis}
\end{figure}

In Figure~\ref{fig:distill-analysis}, we compare the Distillation Decay between Adapters that were initialized using the Low-Shot and Full training data (\algovariant{FL} and \algovariant{FF}, respectively). 
Unsurprisingly, we observe that performing the \textit{Initialize} phase using the full training data results in lower Distillation Decay than only a Low-shot version of the training data (5\%).

\label{subsec:train-analysis}

\subsection{How much does the Final Adapter Training Help?}

In \textit{Phase Three: Train the Adapter}, the Adapter $\Phi_k$ initialized in \textit{Phase Two} is further trained on the training data $\dataset{k}$ of task $\task{k}$. 
We compute the extent to which this training helps by computing the \textit{Average Phase Three Gain}, which is the relative increase in Adapter performance after Phase Three, averaged across tasks $\task{2...K}$ in a given task order. 
In Figure~\ref{fig:train-analysis}, we compare the Average Phase Three Gain for each \algoshort\ variant. 
We observe that \algovariant{LL} benefits the most from the \textit{Phase Three} training, and variants with more exposure to the full training data benefit less.

\section{Conclusions and Future Work}
We propose \algofull\ (\algoshort), a CL algorithm that utilizes knowledge from previously-learned task Adapters to learn an initialization for the incoming task's Adapter module. 
Our experiments demonstrate that \algoshort\ is capable of transferring knowledge between task Adapters for sequences of VQA tasks, outperforming AdapterFusion without incurring the associated large parametric cost. 

There are several opportunities for improving the current design of \algoshort. 
Future work can explore making \algoshort\ more training time efficient---the best-performing variant, \algovariant{FF}, requires three full passes of the training data. 
The \algovariant{LL} variant requires minimal additional training time over normal Adapter training, but struggles to both effectively use existing Adapters in \textit{Phase One: Improvise} and transfer that knowledge to the new task Adapter in \textit{Phase Two: Initialize}. 

Similarly, the \textit{FL} variant suffers from high Distillation Decay in \textit{Phase Two: Initialize} (Section~\ref{subsec:analyze-init}). 
Future work can explore better distillation methods to reduce this decay, so that the initialized Adapters can more effectively use knowledge from other task Adapters learned in \textit{Phase One: Improvise}.

Finally, our method still trains a separate Adapter module for each task. 
This form of model expansion can become costly as the number of tasks scales. 
Future work can explore how distillation-based methods~\citep{ermis2022memory} can mitigate the memory cost of model expansion.

\subsubsection*{Acknowledgments}
This work was supported by the Laboratory for Analytic Sciences (LAS), National Security \& Special Research Initiatives, and in part by DARPA under contract HR001121C0168.

\clearpage

\bibliography{references}
\bibliographystyle{collas2023_conference}

\newpage
\appendix
\onecolumn

\section{ClosestTaskInit Baseline Details}
\label{sec:cti}

We begin by computing pairwise similarity between each of our continual learning tasks. 
Although we can define task similarity using several heuristics, we select the previous task whose inputs are most similar to the new task's inputs.

For each task $\task{k}$, we encode all training examples using the \vilt\ vision-language encoder~\citep{kim2021vilt}, and extract the average representation of the [CLS] token, $h_k$. 
For each pair of tasks $\task{k}$ and $\task{j}$, the task similarity is computed as the cosine similarity between their respective task representation vectors $h_k$ and $h_j$.
\begin{align*}
\text{sim}(\task{k}, \task{j}) = \cos(h_k, h_j)    
\end{align*}
\begin{table*}[t]
    \centering
    \footnotesize
    \begin{tabular}{lrrrrr}
    \toprule
    \multirow{2}{*}{\backslashbox{Task $\task{k}$}{Task $\task{j}$}} & \multicolumn{1}{c}{VQAv2} & \multicolumn{1}{c}{Visual7W} & \multicolumn{1}{c}{VQA-Abs} & \multicolumn{1}{c}{VizWiz} & \multicolumn{1}{c}{DAQUAR}  \\
     & & & & & \\
    \midrule
    VQAv2 & - & 0.9167 & 0.9480 & 0.9877 & 0.9816 \\
    Visual7W & 0.9167 & - & 0.9771 & 0.8670 & 0.9280 \\
    VQA-Abs & 0.9480 & 0.9771 & - & 0.9070 & 0.9541 \\
    VizWiz & 0.9877 & 0.8670 & 0.9070 & - & 0.9673 \\
    DAQUAR & 0.9816 & 0.9820 & 0.9541 & 0.9673 & -\\
    \bottomrule
    \end{tabular}
    \caption{Task similarity score $\text{sim}(h_k, h_j)$ between pairs of tasks $\task{k}$ and $\task{j}$.}
    \label{tab:task-sims}
\end{table*}
The task similarity scores for our 5 visual question answering tasks are presented in Table~\ref{tab:task-sims}. 
For each newly arriving task $\task{k} (k > 1)$, we compute its representation $h_k$, and then find the most similar task $\task{k^*}$.
\begin{align*}
    k^* = \argmax_{j \in \{1...k-1\}} \text{sim}(\task{k}, \task{j})
\end{align*}

We initialize the new Adapter $\Phi_k$ by copying the parameters of Adapter $\Phi_{k^*}$, and also copy the task-specific parameters $\Psi_{k^*}$. 
We then further train the Adapter $\Phi_k$ and task-specific parameters $\Phi_k$ on task $\task{k}$.


\end{document}